%% file: sarvagya.tex
\documentclass[12pt]{article}

\input preamble.tex

\title{Robustness of Selected Learning Models under Label-Flipping Attack}

\author{Sarvagya Bhargava\footnotemark[1]\ \ \ 
Mark Stamp\footnotemark[1]\,\,\footnotemark[2]}

\begin{document}

\symbolfootnotetext[1]{Department of Computer Science, San Jose State University}
\symbolfootnotetext[2]{mark.stamp$@$sjsu.edu}

\maketitle

\abstract
In this paper we compare traditional machine learning and deep learning models trained on a malware dataset 
when subjected to adversarial attack based on label-flipping. Specifically, we investigate the robustness of 
Support Vector Machines (SVM),
Random Forest,
Gaussian \Naive\ Bayes (GNB),
Gradient Boosting Machine (GBM),
LightGBM,
XGBoost,
Multilayer Perceptron (MLP),
Convolutional Neural Network (CNN),
MobileNet,
and
DenseNet models
when facing varying percentages of misleading labels.
We empirically assess the the accuracy of each of these models under 
such an adversarial attack on the training data. 
This research aims to provide insights into which models are inherently
more robust, in the sense of being better able to resist intentional disruptions 
to the training data. We find wide variation in the robustness of the models tested
to adversarial attack, with our MLP model achieving the best combination of initial accuracy
and robustness.

\section{Introduction}

Malicious software---malware---is a pernicious threat. Machine learning models have proven to 
be powerful tools for identifying and mitigating 
malware-based attacks. Since malware evolves, we need to constantly improve our defenses,
which implies that research into learning models as applied in to the 
malware problem is essential.

One of the fundamental areas where we need to improve our defenses is in dealing with adversarial attacks
on machine learning models. Poisoning attacks typically involve corrupting the training data or features
vectors. The research in this paper, focuses on label-flipping adversarial 
attacks~\cite{xiao2015}. These attacks involve mislabeling data points during training, 
which serves to corrupt the training phase, and thereby degrade model performance. Understanding 
how various models respond to these attacks is the main focus of this paper. We consider both classic
machine learning techniques and deep learning models.

Evaluating machine learning models against label-flipping attacks within the malware domain 
is important for the following reasons.
\begin{itemize}
\item The consequences of misclassification in malware detection can be severe, leading to security breaches, 
data compromise, and system vulnerabilities. Thus it is important to understand how different models respond to 
adversarial attacks.
\item Many types of learning models have been shown to perform well in the malware domain. Comparing and evaluating 
the resilience and robustness of these architectures offers critical insights that can guiding practitioners in selecting the 
most suitable models for defensive applications.
\end{itemize}

In short, understanding and mitigating the impacts of label-flipping adversarial attacks is 
imperative for the development of secure, reliable, and effective machine learning based malware detection systems. 
This research advances knowledge in the field by serving as a practical guide for practitioners 
to select and implement more secure machine learning models.

In this paper, we utilize the Malicia dataset (consisting of Windows malware)
to evaluate the resilience of various machine learning and deep learning 
algorithms when faced with label-flipping attacks. Initially, we pre-process data comprising of~11,688 malware binaries, 
which are classified into~48 distinct malware families~\cite{mehta2024natural}. We exclude from our training and testing
all classes containing fewer than~50 samples. We partition the resulting dataset into training and testing subsets, and we implement 
a procedure to simulate label-flipping attacks on the test set. This manipulated dataset is subsequently 
fed into a variety of trained models to assess their performance. These results enable us to analyze 
the effectiveness of the models under this attack scenario. We empirically analyze
the robustness of
Support Vector Machines (SVM),
Random Forest,
Gaussian \Naive\ Bayes (GNB),
Gradient Boosting Machine (GBM),
LightGBM,
XGBoost,
Multilayer Perceptron (MLP),
Convolutional Neural Network (CNN),
MobileNet,
and
DenseNet models.

The remainder of this paper is organized as follows.
In Section~\ref{chap:Related work}, we provide information on related work, that is, selected prior research into 
adversarial attacks involving malware datasets. Section~\ref{chap:Background} covers the technical details of 
our research, including an overview of the machine learning models used in this study. In Section~\ref{chap:experimentation}, 
we detail the experiments conducted to evaluate the resilience of our models against label-flipping attacks. 
The discussion extends to the implications of our findings, emphasizing the strengths and limitations of 
current approaches. We conclude the paper in Section~\ref{chap:future work}, where we also consider 
future work that could be undertaken to extend the results in this paper.

\section{Related work}\label{chap:Related work}

Adversarial attacks against malware detection systems have emerged as a challenging problem 
in cybersecurity research. In this section, we discuss representative examples of previous works 
related to adversarial attacks on malware detection and classification systems.

Aryal et al. in~\cite{aryal2021survey} provide a detailed survey of adversarial attacks within malware 
detection systems. Their work systematically highlights the vulnerabilities of various machine learning 
models to these such threats. Our research aims to build upon this previous work by investigating 
the resiliency of various machine learning and deep learning techniques to label-flipping attacks.
A goal of our research is to uncover any inherent model-specific strengths and weaknesses.

Paudice et al. in~\cite{paudice} conducts an in-depth study utilizing three distinct datasets
(MNIST, BreastCancer, and SpamBase) to explore the efficacy of label-flipping attacks on 
machine learning models. Their research demonstrates the significant impact of 
such adversarial tactics on the performance of learning systems, and they also consider 
a $k$-Nearest Neighbor based defense mechanism. This mechanism focuses on label sanitization, 
effectively identifying and correcting maliciously altered labels to mitigate the adverse effects of these attacks. 

In their research, Xiao, et al. in~\cite{xiao2015} examined the resilience of Support Vector Machines (SVMs) 
against adversarial label noise attacks. Such attacks aim to manipulate SVM classification through strategic 
label-flipping. Their analysis, focuses on both linear and non-linear SVMs, across synthetic and real-world datasets. 

Taheri et al. in~\cite{taheri2020defending} introduce two novel defense strategies against silhouette clustering-based 
label-flipping attacks, specifically designed for deep-learning-based malware systems. Additionally, 
Bootkrajang and Kab\'{a}n in~\cite{Bootkrajang2012} discuss the utility of robust logistic regression algorithms 
that can withstand label-flipping, underscoring the relevance in practical applications.

Aryal et al.in~\cite{10020528} examine the resilience of various machine learning models to label-poisoning 
within the realm of malware detection by evaluating the detrimental impact of data corruption on the performance 
of ML-based malware detectors. This paper emphasize the critical importance of developing robust defense 
mechanisms to safeguard machine learning applications from adversarial attacks.

Jha et al. in~\cite{jha2023label} introduced ``FLIP,'' a novel label-only backdoor attack method that subverts 
machine learning models by manipulating the labels on training data. Demonstrating significant efficacy, 
FLIP achieved a high attack success rate on the CIFAR-10 dataset with a minimal amount of label corruption, 
while maintaining high accuracy on clean data. This highlights a critical vulnerability in machine learning systems 
and underscores the need to understand which models are more susceptible to these types of attacks.

\section{Background}\label{chap:Background}

In this section, we introduce the various learning models that are considered in
our experiments. These models range from classic machine learning techniques to
cutting-edge pre-trained deep learning models.

\subsection{Classic Models}

Support Vector Machines (SVM)~\cite{boswell2002introduction} are powerful supervised learning 
models used for classification and regression tasks. When training an SVM for binary classification, the goal is to find 
a separating hyperplane that splits the classes. SVMs are effective in high-dimensional spaces and can handle 
non-linear relationships via kernel functions. SVMs easily generalize to the multiclass case, where they
are sometimes referred to as Support Vector Classifiers (SVC).


Random Forest~\cite{breiman2001random} models are constructed by using multiple decision trees. 
They are a category of ensemble learning models and often perform well in classification and regression 
tasks. By combining a number of decision trees, a Random Forest reduces overfitting and increases 
the robustness of the model. They are noted for handling high-dimensional data well.


Gaussian \Naive\ Bayes (GNB)~\cite{hand2001idiot} is a probabilistic algorithm which is relatively simple, 
efficient, and can be highly effective in some cases. GNB is a variant of \Naive\ Bayes that 
works especially well when the independence assumption holds true.

\subsection{Boosting Models}\label{sect:boostModels}

We place an emphasis on boosting models, since mislabeled training data is considered
a weakness of boosting~\cite{Stamp_2022}. Thus, we expect that boosting models will 
generally be susceptible to failure under a label-flipping attack, and we would like to determine
whether there are meaningful differences in the robustness of different boosting techniques.

Gradient Boosting Machines (GBM)~\cite{friedman2001greedy} are a class of ensemble learning techniques 
which are known for incrementally improving model accuracy. This is achieved by generating new models to 
correct misjudgments of preceding models. These models are generated in sequence until no substantial 
improvements are observable. GBM employs decision trees as the base learners and refines them 
through an iterative approach. Specifically, GBM minimizes a loss function by employing weak learners, 
following a method akin to gradient descent. This process addresses errors primarily by focusing on 
the residuals of earlier learners in the sequence, and is accomplished through the sequential addition of 
shallow trees tailored to correct previous mistakes.


LightGBM~\cite{ke2017lightgbm} is a gradient boosting ensemble modeling technique, which focuses on fast
and efficient training with reduced memory usage. LightGBM uses a histogram-based method where it bins 
the data using a histograms of the distribution which, in turn are used to iterate, calculate the gain, and split the data. 
LightGBM also uses feature bundling, where it combines various features together to reduce dimensionality 
and make the training more efficient.


XGBoost~\cite{chen2016xgboost} (eXtreme Gradient Boosting) is an enhancement to the foundational concepts 
of GBM. The benefits of XGBoost are that it is efficient to train, it handles complex relationships, 
it employs regularization techniques that reduce overfitting, it can incorporate parallel processing to improve 
computation speed, and it is robust.

\subsection{Deep Learning Models}

Multilayer Perceptrons (MLP)~\cite{rumelhart1986learning} are a type of feedforward artificial neural network 
characterized by multiple layers of interconnected nodes (i.e., neurons). An MLP has an input layer and 
an output layers, along with one or more hidden layers, with each layer being full-connected to the layers 
above and below. 
MLPs often perform well even on relatively small datasets.


Convolutional Neural Networks (CNN)~\cite{lecun1998gradient} are a category of deep learning algorithms 
that are designed to be efficient for dealing with data where local structure dominates, such as is the case for
images. The architecture 
of a CNN typically involves a sequence of interleaved convolutional and pooling layers, with one or more 
fully connected layers for classification. The convolutional layers apply a number of filters to the input to 
create feature maps that abstract higher-level features from the raw input data. Pooling layers reduce the
dimensionality for the next convolutional layer, thereby reducing the number of parameters and improving the
computational efficiency. CNNs have proven to be highly effective for image image classification 
and object detection, and have been successfully applied to many non-image problems as well.


MobileNets~\cite{howard2017mobilenets} are a streamlined class of convolutional neural networks 
designed for efficiency and are suitable for environments with limited computational resources such as mobile devices. 
MobileNets employ a unique architecture involving depthwise separable convolutions, significantly reducing 
the number of parameters and computational overhead. This makes MobileNets particularly suitable for small datasets, 
as their compact structure minimizes the risk of overfitting while facilitating faster training via transfer learning. 


DensetNets~\cite{huang2017densely} have shown remarkable performance in image classification, 
object detection, and segmentation tasks. Their ability to leverage information from 
previous layers makes them particularly effective for tasks where preserving spatial hierarchies in images is 
crucial.

\section{Experiments and Results}\label{chap:experimentation}

This section provides details on all of our label-flipping experiments. We begin with 
a discussion of the dataset, the preprocessing of the data, and feature extraction. 
We then move on to the experimental results for each of the models, 
where we vary the percentages of labels that are flipped.

\subsection{Dataset and Data Preprocessing}

We train models using the Malicia dataset~\cite{nappa2015malicia}.
In the preprocessing phase, the dataset is filtered to seven malware families, 
based on the criterion that each family should have at least~50 samples. 
The malware families and number of samples per family
are listed in Table~\ref{tab:malware_samples}. In all of our experiments, 
we consider multiclass classification, based on the seven classes in 
Table~\ref{tab:malware_samples}.

\begin{table}[!htb]
\centering\def\z{\phantom{0}}
\caption{Number of samples}\label{tab:malware_samples}
\adjustbox{scale=1.0}{
\begin{tabular}{cc}
\toprule
\textbf{Family} & \textbf{Samples} \\ \midrule
Cridex & \z\z74 \\ 
Harebot & \z\z53 \\ 
SecurityShield & \z\z58 \\ 
Smarthdd & \z\z68 \\ 
Winwebsec & 4360 \\ 
Zbot & 2136 \\ 
Zeroaccess & 1305 \\ \midrule
\textbf{Total} & 8054 \\ \bottomrule
\end{tabular}
}
\end{table}

The models introduced in Section~\ref{chap:Background} 
can be categorized as follows.
\begin{itemize}
\item Classic models --- SVM, Random Forest, and GNB
\item Boosting models --- GBM, Light GBM, and XGBoost 
\item Deep learning models --- MLP, CNN, MobileNet, and DenseNet
\end{itemize}
Initially, we train each model without any label-flipping. 
Then we test each of these model by varying the percentage of labels randomly 
flipped during training, and we discuss the insights gained from these experiments.
The percentage of labels flipped ranges from~10\%\ to~100\%, in increments of~10\%. 
Note that the smallest class, Harebot, has only~53 samples, while the largest class, Winwebsec, 
has~4360 samples. Label-flipping is implemented on a per-class basis, 
that is, for a given flipping percentage, that percentage of labels is randomly flipped in the
training data for each class. 


To train our classic machine learning, boosting models, and MLP, features are obtained by
extracting the mnemonic opcodes, and applying the
TF-IDF vectorizer to the sequence extracted from each sample. 
This method was chosen because TF-IDF is effective at emphasizing crucial information 
within a sequence, while also serving to minimize background noise. 

For training our CNN and pre-trained deep learning models, 
a different preprocessing approach was necessary,
since these models expect image data.
To accommodate this case, we convert each malware sample into an image representation 
by assigning a unique number to each opcode and interpreting the 
first~4096 opcodes as a~$64\times 64$ image. If a sample has fewer 
than~4096 opcodes, we simply pad with~0 to fill out the~$64\times 64$ image.

\subsection{Baseline Results}

\begin{figure}[!htb]
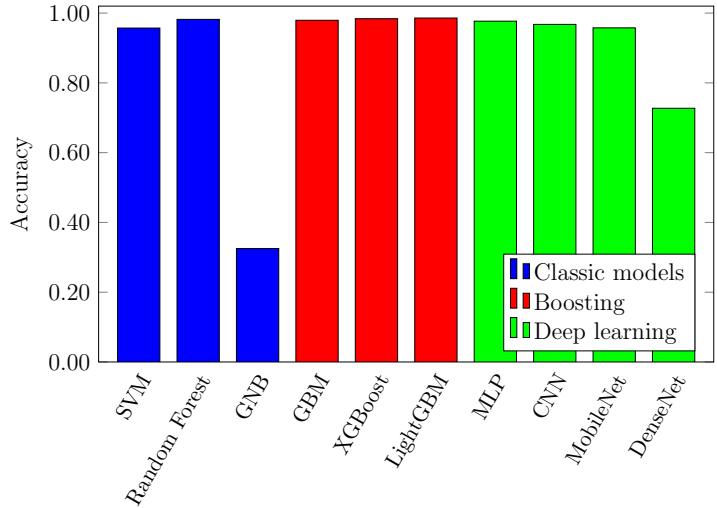

    \centering
    \input figures/bar_init.tex
    \caption{Baseline accuracies without label-flipping}\label{fig:init}
\end{figure}

First, we train each of the~10 models under consideration
on clean data, that is, data without any label-flipping.
These results are summarized in the form of a bar graph in
Figure~\ref{fig:init}. Here, accuracy is defined as 
the number of correctly classified samples divided by the
total number of samples classified.

From Figure~\ref{fig:init}, we observe that a eight of the~10 models perform 
well, with the
top five models (Random Forest, GBM, XGBoost, LightGBM, MLP)
all achieving about~98\%\ accuracy, or higher.
The next three best (SVM, CNN, MobileNet) all 
attain an accuracy of about~96\%.
Only the DenseNet and GNB models fail to produce strong
results on this dataset.

The differences in accuracy among the top eight models is relatively small.
Hence, we might be willing to choose from among these models 
based on robust their inherent robustness to
label-flipping attack, as opposed to accuracy alone. 
Next, we consider label-flipping attacks on each of the~10 models.

\subsection{Label-Flipping Results for Classic Models}

As discussed above, the traditional machine learning models we selected for our experimentation 
are SVM, Random Forest, and GNB. Each of these models was chosen for its distinct approach to data analysis: 
SVM excels in separating data in high-dimensional spaces through margin maximization, 
Random Forest leverages ensembles of decision trees to improve predictive accuracy and robustness, 
and GNB relies on the probabilistic assumptions of data distributions.
Here, we present and discuss the results of our label-flipping experiment for each of these models.
 
\subsubsection{Support Vector Machine Results}

From Figure~\ref{fig:classic}(a) we observe that SVM achieved high accuracy and that
the accuracy was virtually unchanged until more than~60\%\ of the labels were flipped,
and even at~70\%\ label-flipping, the accuracy only diminished slightly. After~70\%\ label
flipping, the accuracy drops precipitously.
These results indicate that SVM is remarkably robust when faced with a label-flipping
adversarial attack.

\subsubsection{Random Forest Results}

In Figure~\ref{fig:classic}(b) we see that the accuracy of our Random Forest model is very high without any
label-flipping. The accuracy then degrades consistently, and almost linearly up to about~60\%\ label-flipping.
Although Random Forest is the most accurate of our classic models, it is not as robust to label-flipping
attacks as SVM (and MLP, as we note below).

\subsubsection{Gaussian \Naive\ Bayes Results}

\begin{figure}[!htb]
\centering
\begin{tabular}{cc}
\includegraphics[width=0.4125\textwidth]{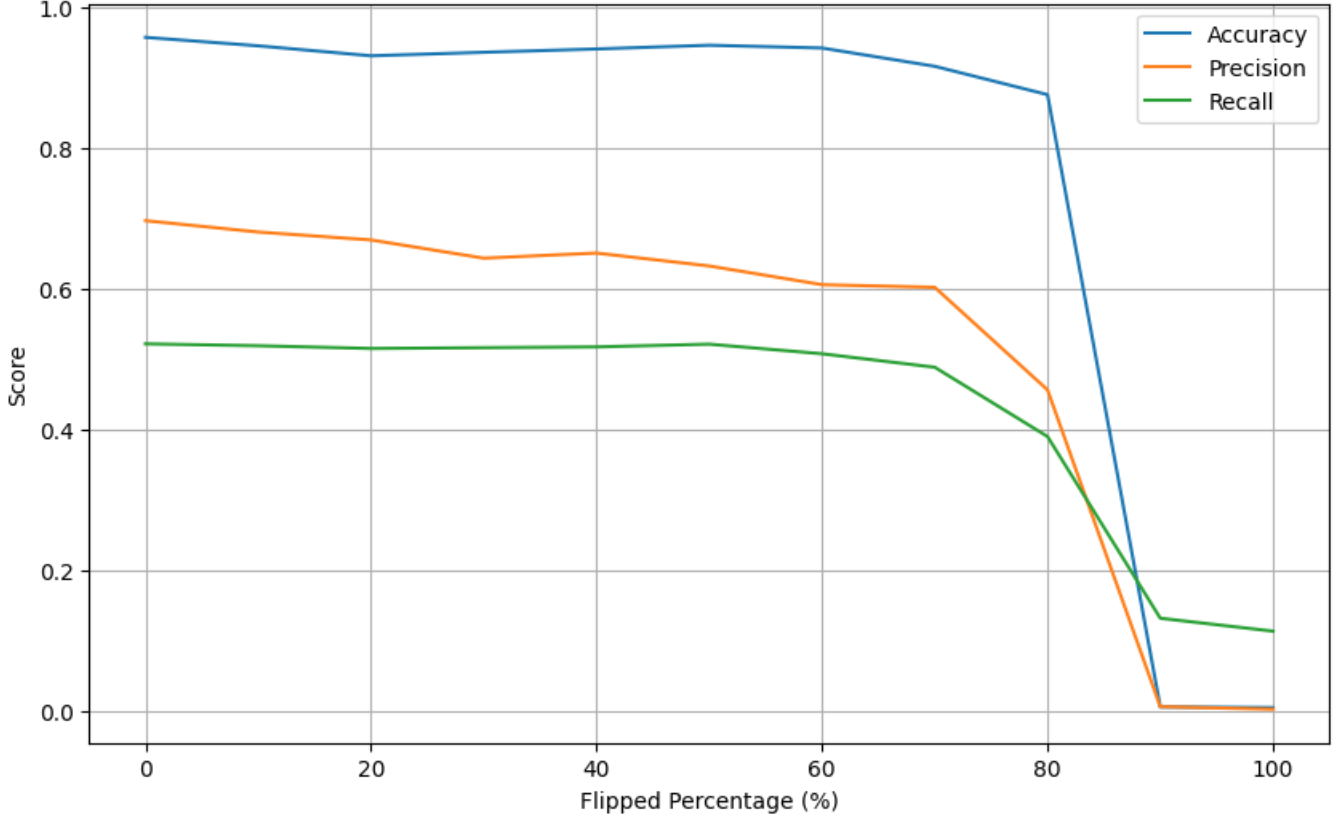}
&
\includegraphics[width=0.4125\textwidth]{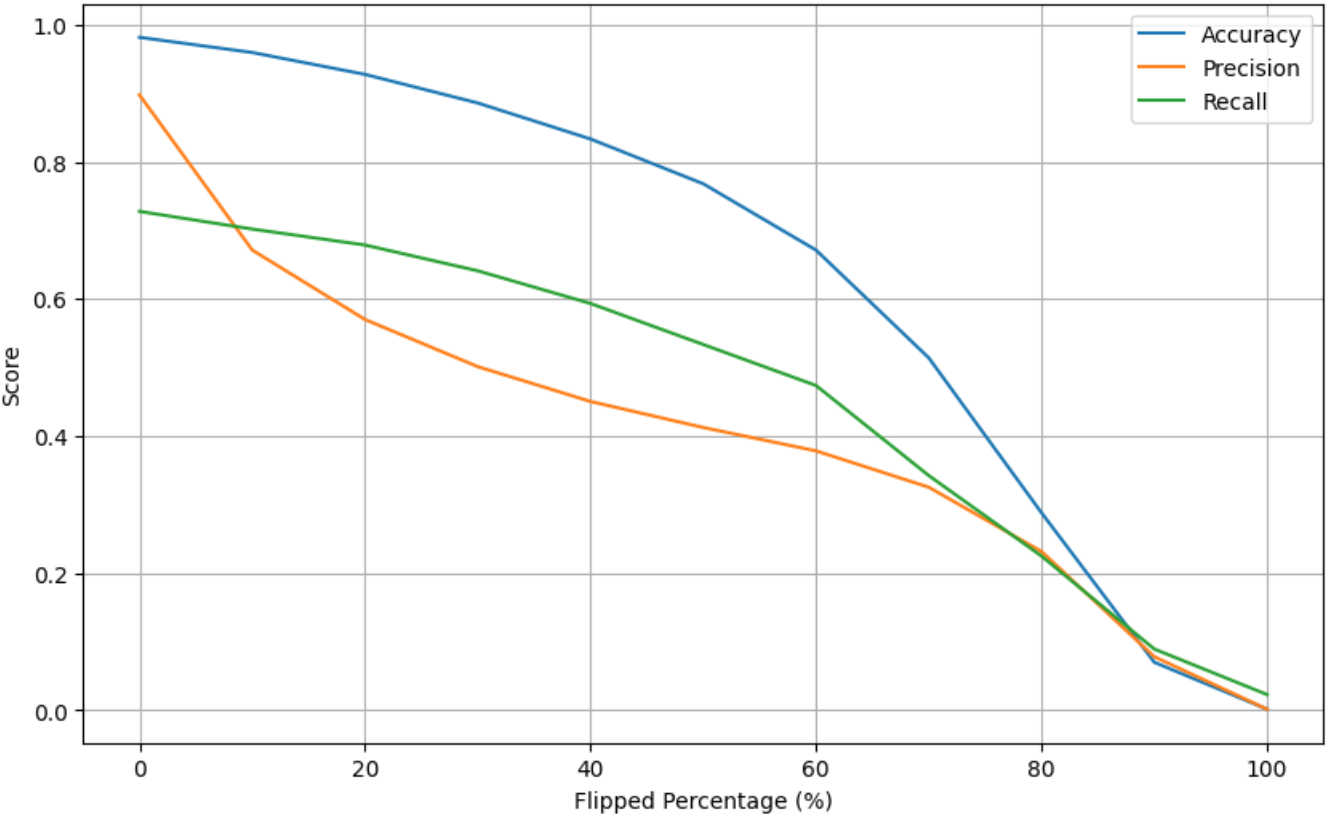}
\\
\adjustbox{scale=0.85}{(a) SVM}
&
\adjustbox{scale=0.85}{(b) Random Forest}
\\
\\[-1.0ex]
\multicolumn{2}{c}{\includegraphics[width=0.4125\textwidth]{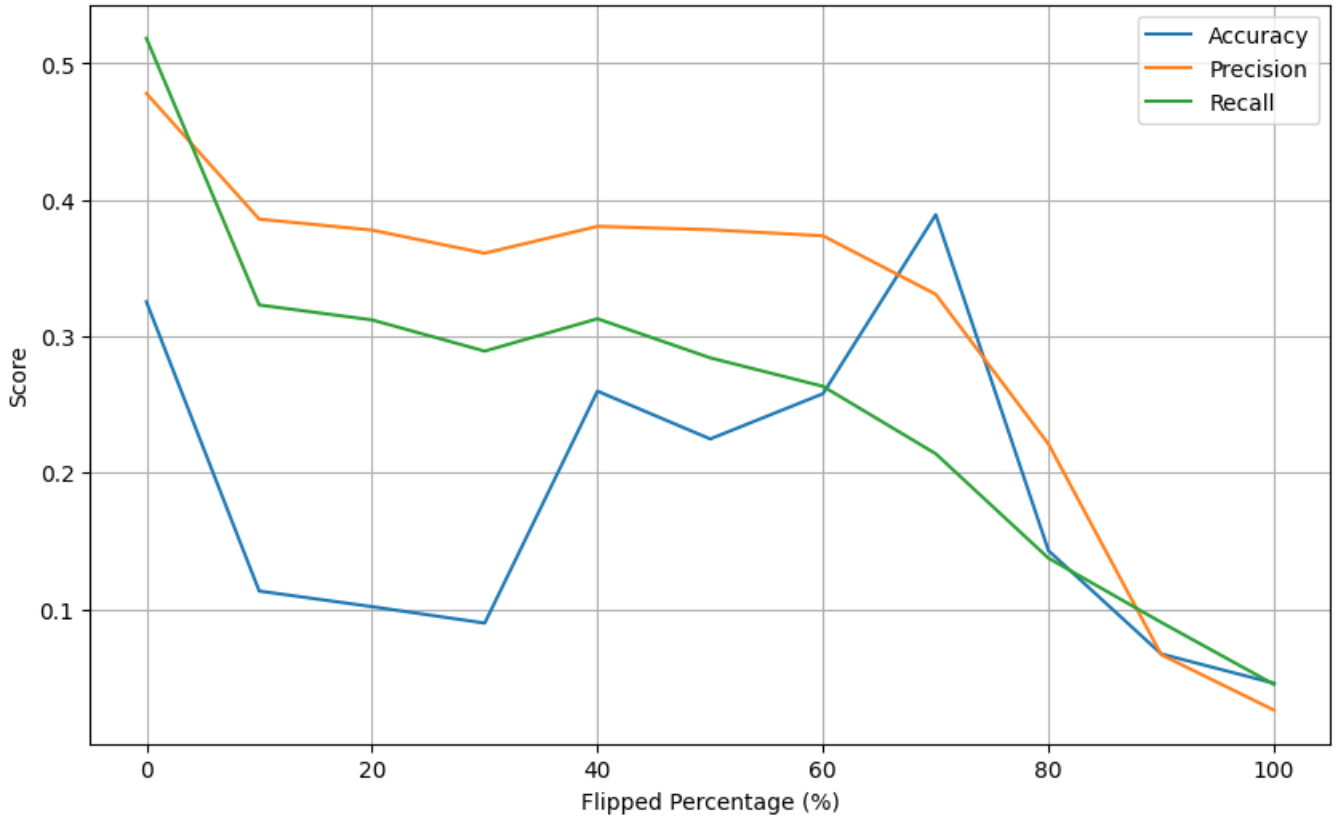}}
\\
\multicolumn{2}{c}{\adjustbox{scale=0.85}{(c) GNB}}
\end{tabular}
\caption{Accuracy, precision and recall graphs for classic ML techniques}\label{fig:classic}
\end{figure}

Figure~\ref{fig:classic}(c) shows that GNB performed very poorly initially and, of course, it also
performed poorly with respect to label-flipping. This model is clearly not suitable for this particular
problem, most likely due to the selected features failing to be conditionally independent.
 
\subsection{Label-Flipping Results for Boosting Techniques}

We also consider label-flipping attacks on advanced boosting techniques. As discussed above,
the specific models we consider are XGBoost, GBM, and LightGBM

\subsubsection{Gradient Boosting Machine Results}

Our GBM results appear in Figure~\ref{fig:boost}(a).We see that this model delivers strong performance
and robustness to label-flipping adversarial attack.
The results for GBM are comparable to the MLP model in Figure~\ref{fig:deep}(a), below.

\subsubsection{XGBoost Results}

From the Figure~\ref{fig:boost}(b), we observe that qualitatively,
XGBoost performs similarly to the Random Forest model in Figure~\ref{fig:classic}(b),
with XGBoost is slightly more robust to label-flipping.
This result is not too surprising, since XGBoost and Random Forest are both based on multiple decision trees.
It is also worth noting that XGBoost has similar initial accuracy as GBM, but it is far less robust
in the face of label-flipping.

\subsubsection{LightGBM Results}

\begin{figure}[!htb]
\centering
\begin{tabular}{cc}
\includegraphics[width=0.4125\textwidth]{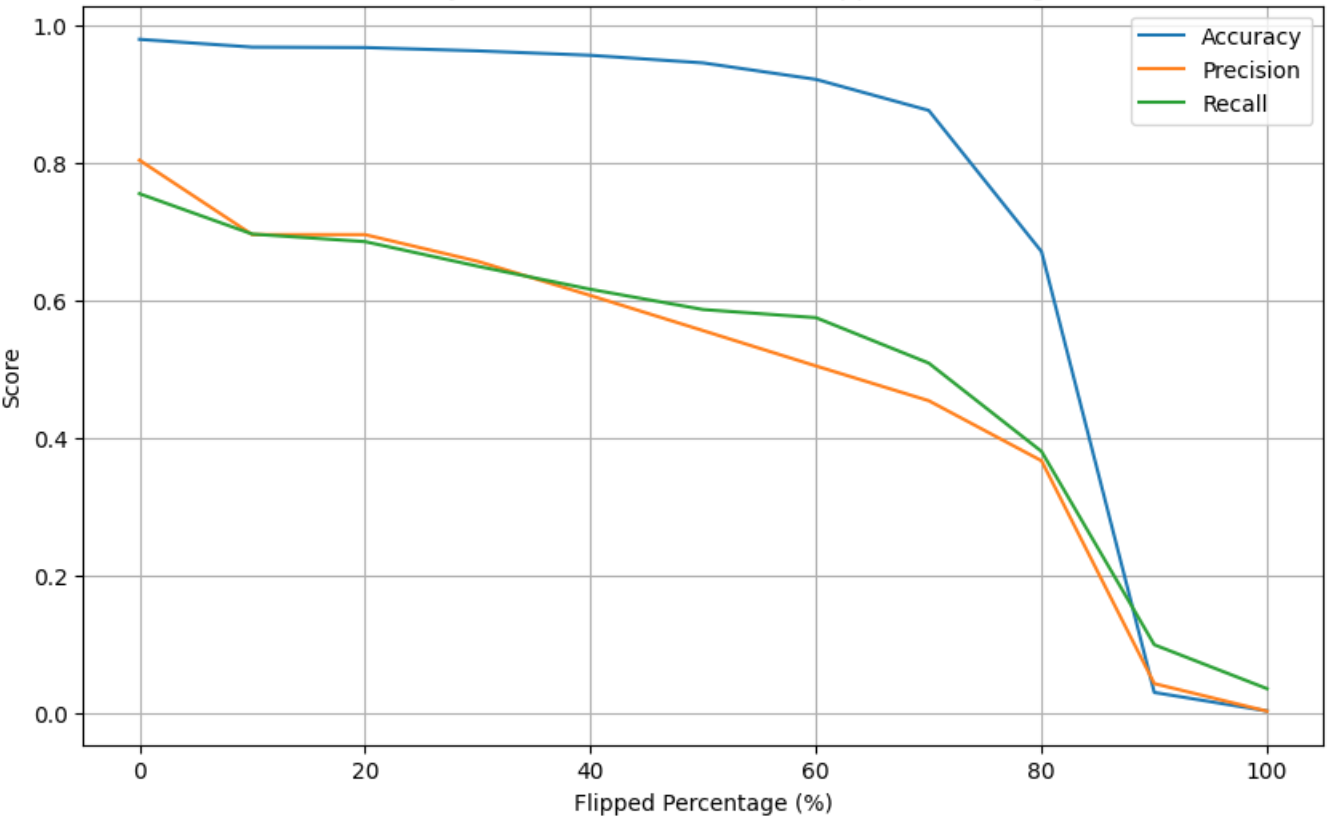}
&
\includegraphics[width=0.4125\textwidth]{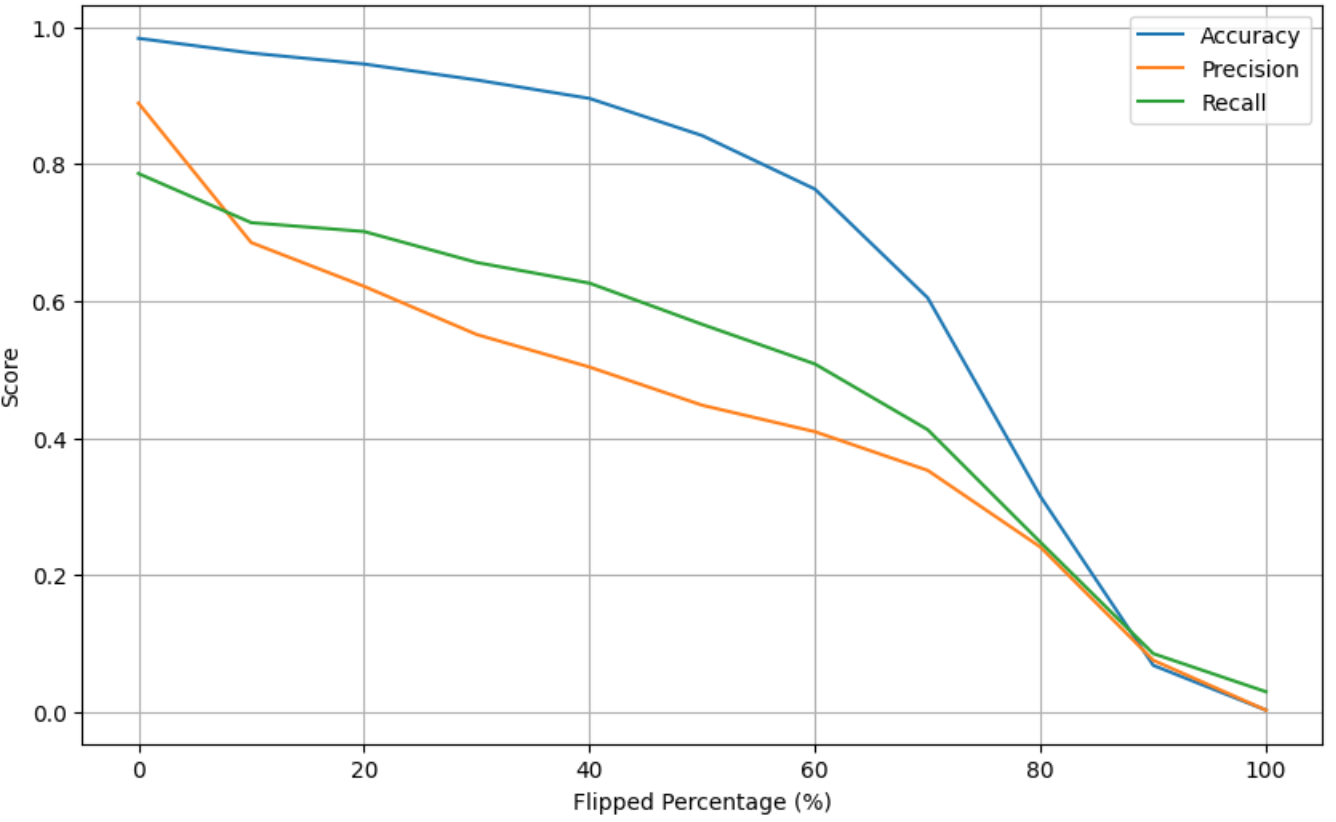}
\\
\adjustbox{scale=0.85}{(a) GBM}
&
\adjustbox{scale=0.85}{(b) XGBoost}
\\
\\[-1ex]
\multicolumn{2}{c}{\includegraphics[width=0.4125\textwidth]{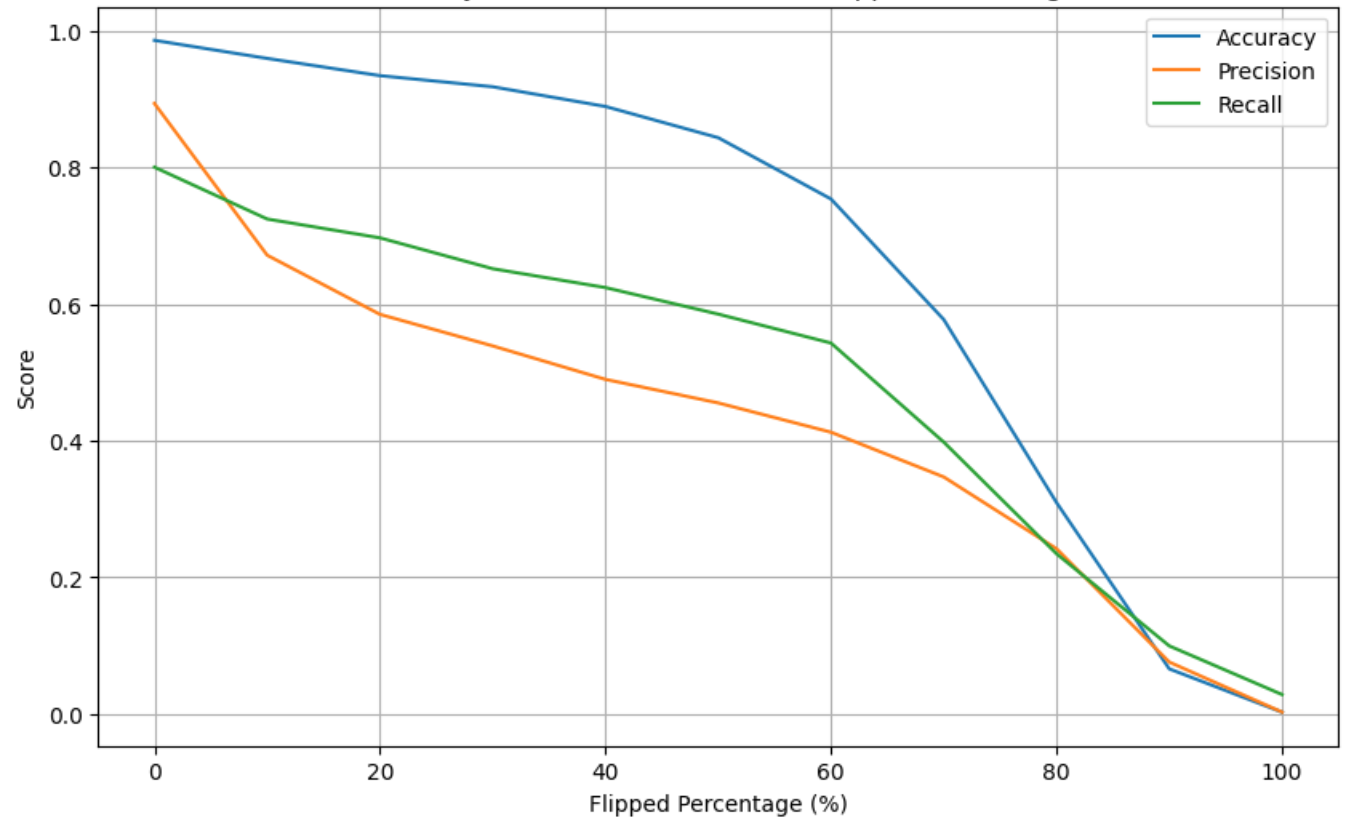}}
\\
\multicolumn{2}{c}{\adjustbox{scale=0.85}{(c) LightGBM}}
\end{tabular}
\caption{Accuracy, precision and recall graphs for boosting techniques}\label{fig:boost}
\end{figure}

In Figure~\ref{fig:boost}(c) we see that LightGBM yields almost identically performance as our XGBoost model,
but well below that of the GBM model.
This is interesting, as it indicates that the LightGBM is---in the sense of robustness---much weaker than
the GBM model from which it is derived.

\subsection{Label-Flipping Results for Deep Learning Models}

In addition to traditional machine learning models and boosting models, 
we consider deep learning architectures. As discussed above, we analyze three image-based
deep learning models, namely, MLP, a generic CNN, as well as the pre-trained models MobileNet and DenseNet.

\subsubsection{Multilayer Perceptron Results}

As can be seen in Figure~\ref{fig:deep}(a), our MLP model performs similar to---although slightly better 
than---the SVM model, both initially, and at each label-flipping percentage. The similarity
of SVM and MLP is not too surprising, as these are closely related 
techniques. Roughly speaking, an MLP can be viewed as a generalization of
an SVM, where the
equivalent of the kernel function is learned, rather than being specified 
as a hyperparameter during training~\cite{Stamp_2022}.

\subsubsection{Convolutional Neural Network Results}

From the graphs in Figure~\ref{fig:deep}(b), we see that our CNN model gives us accuracies comparable 
to the Random Forest model in Figure~\ref{fig:classic}(b). This model is not nearly as robust as
the classic SVM and MLP models, and it also is far weaker than the GBM model.

\subsubsection{MobileNet Results}

In Figure~\ref{fig:deep}(c), we observe that, as compared to CNN, the performance of MobileNet is 
slightly better across the full range of label-flipping attacks. However, as with our CNN model,
MobileNet trails far behind the SVM, MLP, and GBM models.

\subsubsection{DenseNet Results}

\begin{figure}[!htb]
\centering
\begin{tabular}{cc}
\includegraphics[width=0.4125\textwidth]{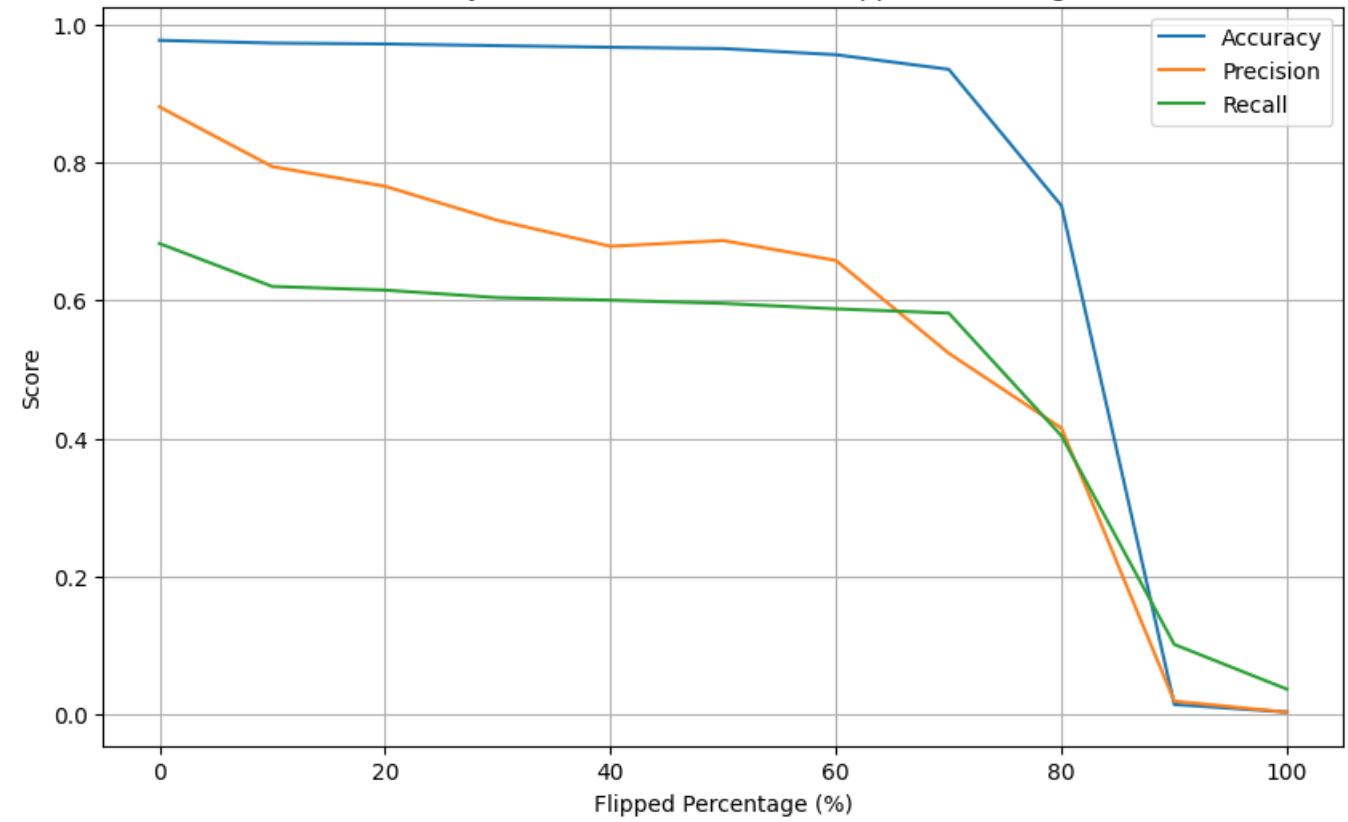}
&
\includegraphics[width=0.4125\textwidth]{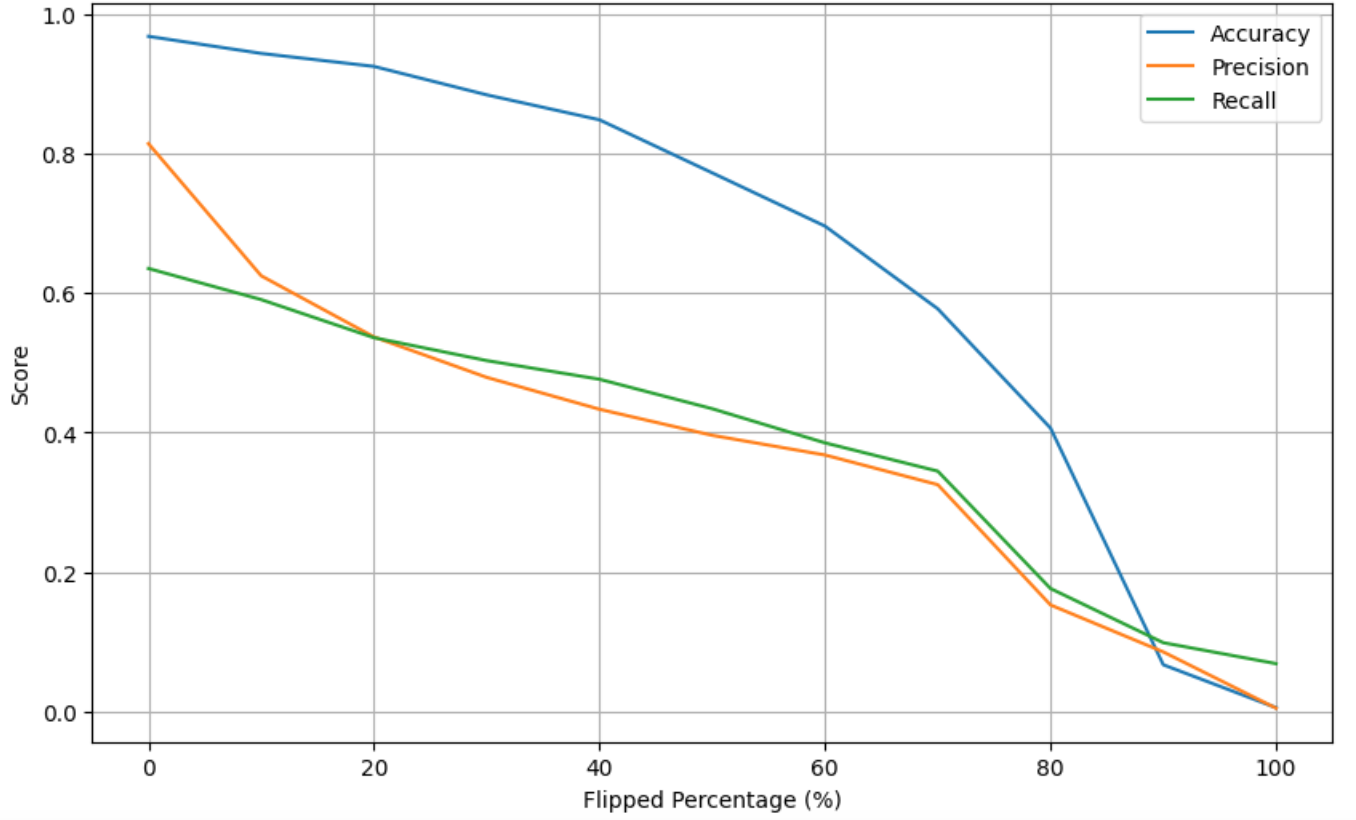}
\\
\adjustbox{scale=0.85}{(a) MLP}
&
\adjustbox{scale=0.85}{(b) CNN}
\\
\\[-1ex]
\includegraphics[width=0.4125\textwidth]{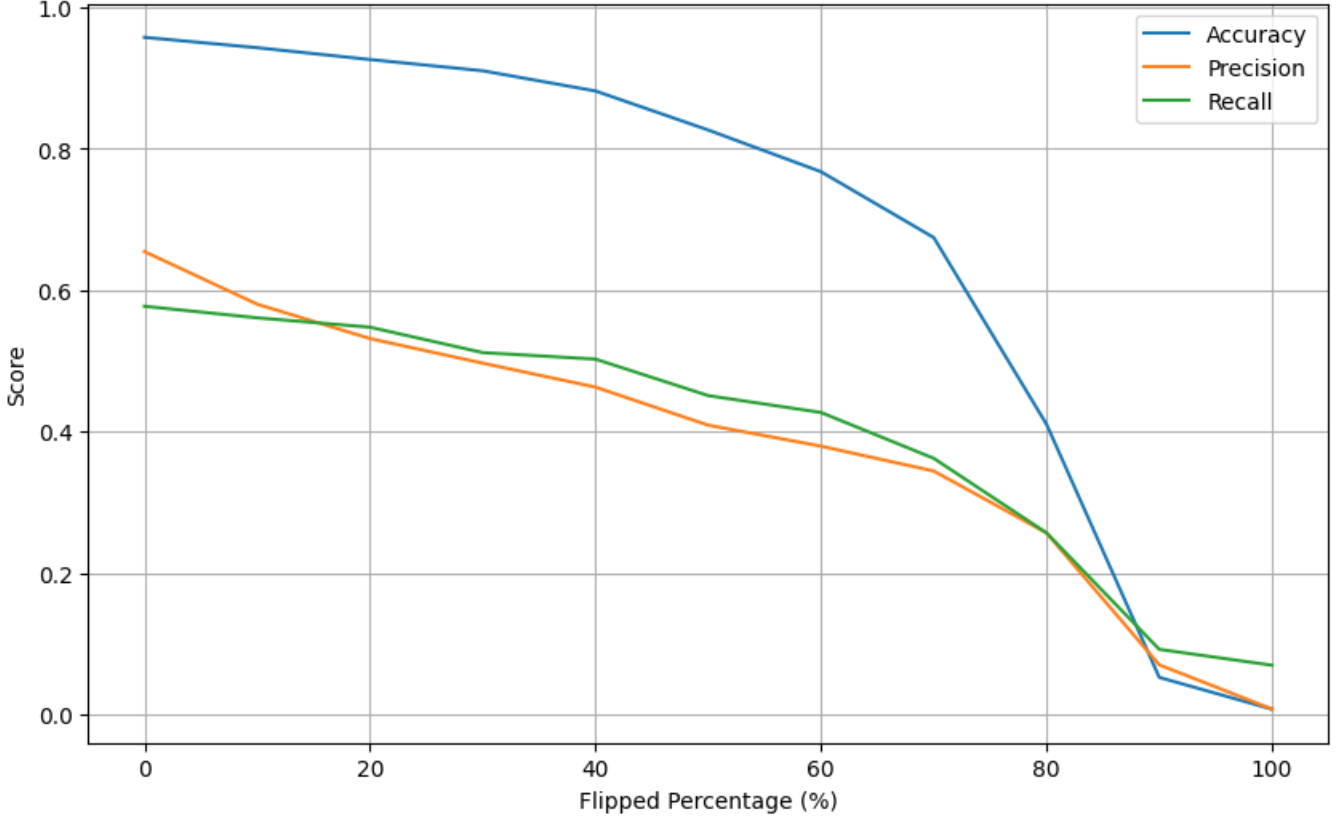}
&
\includegraphics[width=0.4125\textwidth]{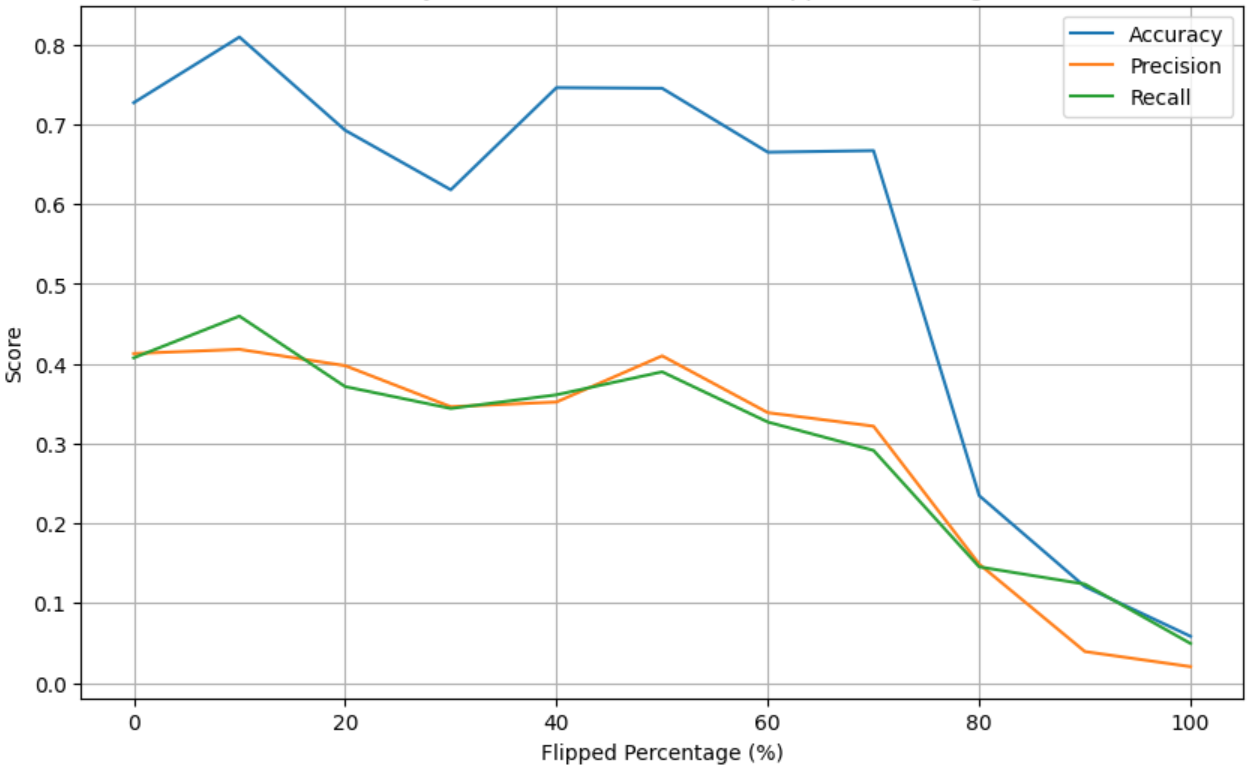}
\\
\adjustbox{scale=0.85}{(c) MobileNet}
&
\adjustbox{scale=0.85}{(d) DenseNet}
\end{tabular}
\caption{Accuracy, precision and recall graphs for deep learning techniques}\label{fig:deep}
\end{figure}

DenseNet results in Figure~\ref{fig:deep}(d). We found DenseNet difficult to train, and hence
the poor and erratic results for this model are not surprising. We believe that there is insufficient
data in our training set for this particular model.

\subsection{Discussion}

\begin{figure}[!htb]
\centering
\begin{tabular}{cc}
\multicolumn{2}{c}{\includegraphics[width=0.5\textwidth, height=0.25\textheight]{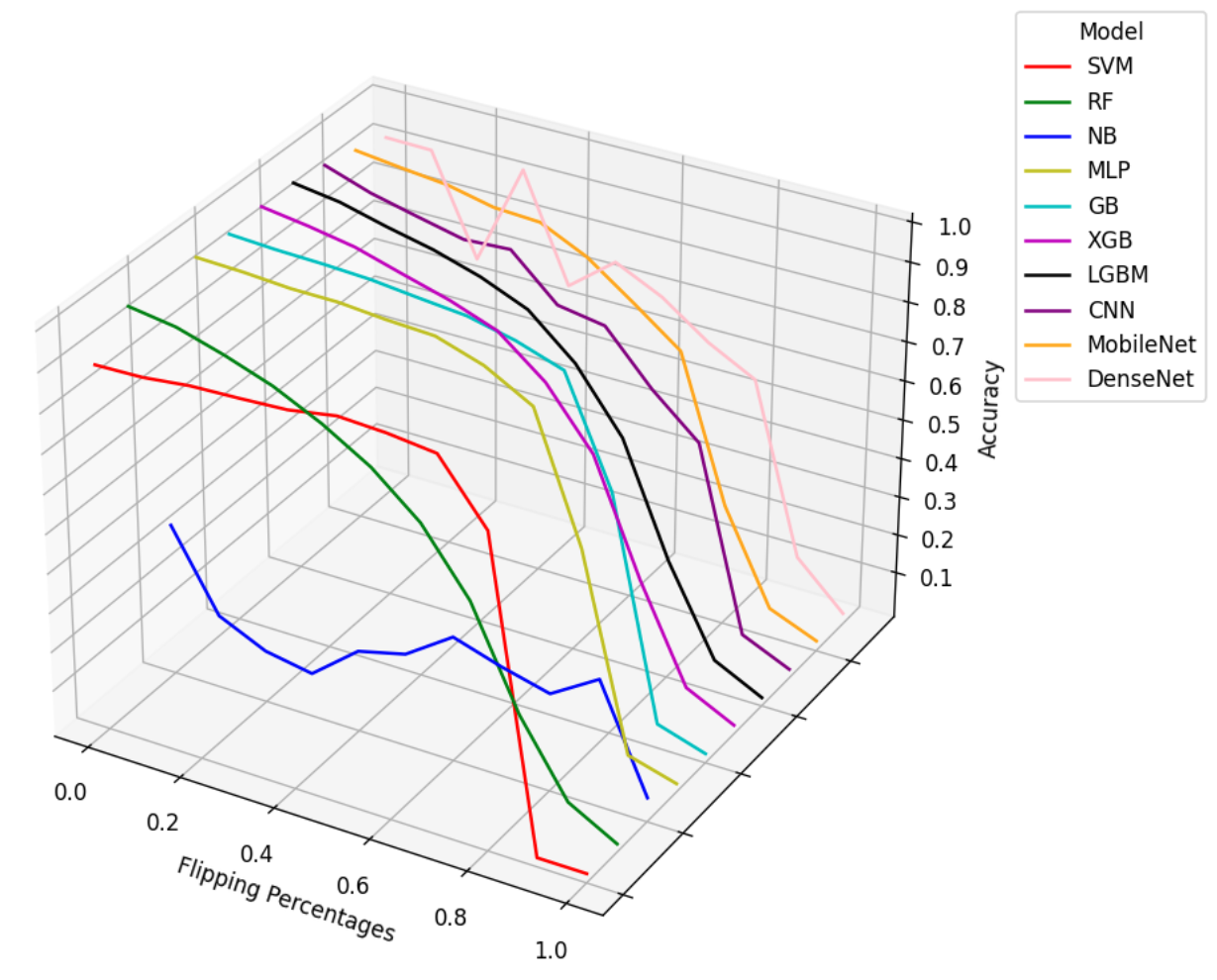}}
\\
\multicolumn{2}{c}{\adjustbox{scale=0.85}{(a) Accuracy}}
\\
\\[-3ex]
\includegraphics[width=0.4125\textwidth]{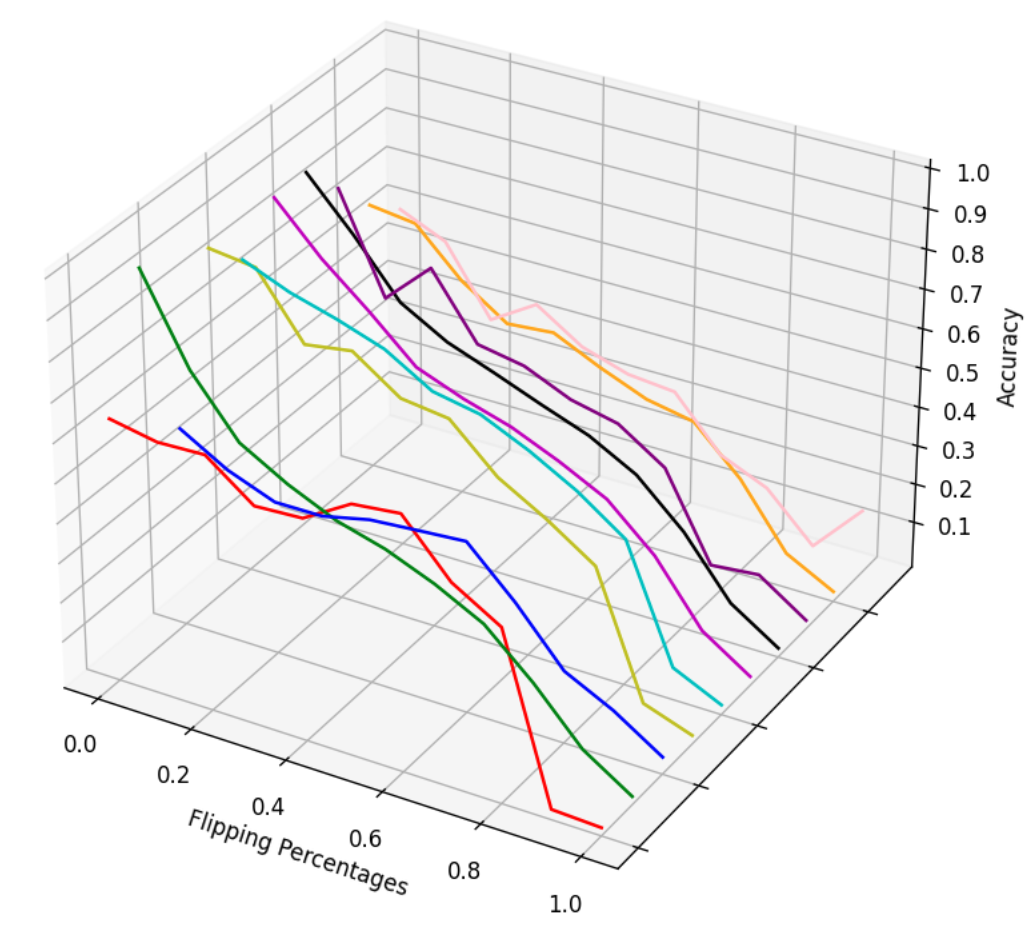}
&
\includegraphics[width=0.4125\textwidth]{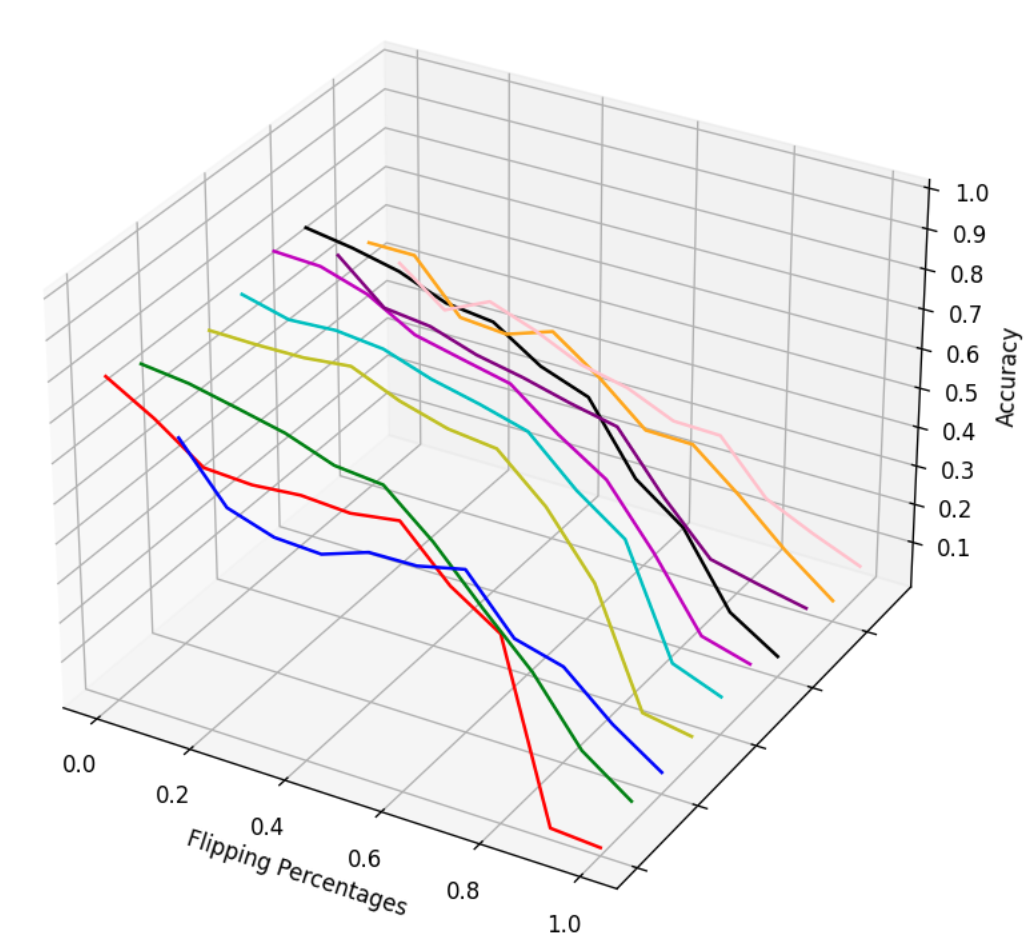}
\\
\adjustbox{scale=0.85}{(b) Precision}
&
\adjustbox{scale=0.85}{(c) Recall}
\end{tabular}
\caption{Accuracy, precision, and recall for all models tested}\label{fig:scores}
\end{figure}

Figure~\ref{fig:scores}(a) depicts the accuracy of all models tested,
while Figures~\ref{fig:scores}(b) and~(c) give the precision and recall, respectively.
These graphs serve to emphasize that, overall, our best model is the MLP.
The MLP has nearly the highest initial accuracy, and 
it is remarkably robust to label-flipped training data. 
The SVM model yields slightly worse results than MLP, while
also providing robustness. The GBM model also performs
well, both in terms of initial accuracy, and robustness to label-flipping.

CNN and MobileNet, two of the three image-based deep learning techniques considered, 
performed well on the malware classification problem. 
However, these two techniques are quite fragile with respect to label
flipping.

\section{Conclusion and Future Work}\label{chap:future work}

In this paper, we compared the robustness of various learning model under a label-flipping attack scenario.
The underlying learning problem was malware classification, and the we considered a variety of
classic machine learning techniques, 
boosting techniques,
and deep learning techniques. Specifically, the classic techniques tested were
Support Vector Machine (SVM), 
Random Forest, 
and Gaussian \Naive\ Bayes (GNB);
the boosting techniques we analyzed were
Gradient Boosting Machine (GBM), 
XGBoost, and LightGBM;
while the deep learning techniques were
Multilayer Perceptron (MLP), 
Convolutional Neural Network (CNN), 
MobileNet, 
and DenseNet.
Although most of these techniques performed well on the original classification problem,
the MLP and SVM were the most robust, with the boosting
technique of GBM also performing well with respect to robustness. 
The Random Forest model was the least robust, while the image-based
models and two of the boosting techniques (XGBoost and LightGBM)
also did not hold up well under our label-flipping adversarial attack.

These results have practical implications. In an environment where adversarial attacks are
likely, and defenses could be challenging to implement, we might be willing to give up a small
amount of initial accuracy for a model that is inherently more robust to such an attack.
Of the models tested, MLP stands out as giving high initial accuracy---within~1\%\ of
the best model---yet also being the most robust under a label-flipping scenario.
Furthermore, as mentioned in Section~\ref{sect:boostModels},
mislabeled data is generally considered to be an inherent weakness of 
boosting techniques, However, we found that GBM is reasonably robust in this regard.
Thus, GBM might be preferred in cases where a boosting strategy is needed, and
mislabeled data (or label-flipping attack) is a legitimate concern.

There are many possible avenues for future work. Additional models could be considered,
as well as additional datasets and learning problems. We could consider more advanced
and targeted label-flipping attacks, as well as other classes of attacks. Defenses against attacks, 
and countermeasures to those defenses would be additional interesting related problems.

\bibliographystyle{plain}
\bibliography{references.bib}

\end{document}

%% file: preamble.tex
\usepackage{amsmath,amsthm, amsfonts, amssymb, amsxtra, amsopn}
\usepackage{pgfplots}
\pgfplotsset{compat=1.17}
\usepackage{pgfplotstable}
\usepackage{graphicx,grffile}
\usepackage{multirow}
\usepackage{algorithm} 
\usepackage[noend]{algpseudocode} 
\usepackage{longtable}
\usepackage{booktabs}
\usepackage{listings}
\usepackage{cmap}
\usepackage{colortbl}
\usepackage{adjustbox}
\usepackage{epsfig}

\usepackage[tableposition=top,font=small,skip=5pt]{caption}
\usepackage{subcaption}
\usepackage{makecell} 

\PassOptionsToPackage{hyphens}{url}

\usepackage{hyperref}
\hypersetup{colorlinks=true,linkcolor=black,citecolor=black,urlcolor=blue,filecolor=black}
\hypersetup{pdfpagemode=UseNone,pdfstartview=}

\usepackage{enumitem}
\setlist[itemize]{noitemsep, topsep=0pt}

\advance\oddsidemargin by -0.4in
\advance\textwidth by 0.8in

\advance\topmargin by -0.4in
\advance\textheight by 0.8in

\long\def\symbolfootnotetext[#1]#2{\begingroup%
\def\thefootnote{\fnsymbol{footnote}}\footnotetext[#1]{#2}\endgroup}

\let\ii=\i

\def\i{{\tt i}}

\def\z{{\tt z}}

\def\Naive{Na\"{\ii}ve}

%% file: figures/bar_init.tex
\begin{tikzpicture}[scale=0.8, every node/.style={scale=0.95}]
\pgfkeys{/pgf/number format/.cd,1000 sep={}}
\begin{axis}[
        width  = 0.75*\textwidth,
        height = 7.5cm,
        ymin=0.0, ymax=1.02,
        ytick={0.0, 0.2, 0.4, 0.6, 0.8, 1.0},
        major x tick style = transparent,
        ybar=5*\pgflinewidth,
        bar width=20.0pt,
        ylabel = {Accuracy},
        symbolic x coords={SVM,Random Forest,GNB,GBM,XGBoost,LightGBM,MLP,CNN,MobileNet,DenseNet},
	y tick label style={
    		/pgf/number format/.cd,
   		fixed,
   		fixed zerofill,
    		precision=2},
        x tick label style={
        		rotate=60,
		font=\small,
		anchor=north east,
		inner sep=0mm
		},
	legend cell align=left,
        legend pos=south east,
        enlarge x limits=0.075,
        every axis plot/.append style={
        		bar shift=0pt,
		fill
        }
]
\addplot [fill=blue,opacity=1.00]
coordinates {
(SVM, 0.9571)
(Random Forest, 0.9820)
(GNB, 0.3253)
};
\addlegendentry{Classic models}
\addplot [fill=red,opacity=1.00]
coordinates {
(GBM, 0.9794)
(XGBoost, 0.9839)
(LightGBM, 0.9858)
};
\addlegendentry{Boosting}
\addplot [fill=green,opacity=1.00]
coordinates {
(MLP, 0.9768)
(CNN, 0.9676)
(MobileNet, 0.9577)
(DenseNet, 0.7272)
};
\addlegendentry{Deep learning}
\end{axis}
\end{tikzpicture}

%% file: sarvagya.bbl
\begin{thebibliography}{10}

\bibitem{aryal2021survey}
Kshitiz Aryal, Maanak Gupta, and Mahmoud Abdelsalam.
\newblock A survey on adversarial attacks for malware analysis.
\newblock \url{https://arxiv.org/abs/2111.08223}, 2021.

\bibitem{10020528}
Kshitiz Aryal, Maanak Gupta, and Mahmoud Abdelsalam.
\newblock Analysis of label-flip poisoning attack on machine learning based
  malware detector.
\newblock In {\em 2022 IEEE International Conference on Big Data}, pages
  4236--4245, 2022.

\bibitem{Bootkrajang2012}
Jakramate Bootkrajang and Ata Kab\'{a}n.
\newblock Label-noise robust logistic regression and its applications.
\newblock In Peter~A. Flach, Tijl De~Bie, and Nello Cristianini, editors, {\em
  Machine Learning and Knowledge Discovery in Databases}, ECML PKDD, 2012.

\bibitem{boswell2002introduction}
Dustin Boswell.
\newblock Introduction to support vector machines.
\newblock
  \url{https://www.semanticscholar.org/paper/Introduction-to-Support-Vector-Machines-Boswell/ea2ea7c6e280c1cfb67ee38ea63a327b1ba3ca36},
  2002.

\bibitem{breiman2001random}
Leo Breiman.
\newblock Random forests.
\newblock {\em Machine Learning}, 45:5--32, 2001.

\bibitem{chen2016xgboost}
Tianqi Chen and Carlos Guestrin.
\newblock {XGBoost}: A scalable tree boosting system.
\newblock In {\em Proceedings of the 22nd ACM SIGKDD International Conference
  on Knowledge Discovery and Data Mining}, pages 785--794, 2016.

\bibitem{friedman2001greedy}
Jerome~H. Friedman.
\newblock Greedy function approximation: A gradient boosting machine.
\newblock {\em Annals of Statistics}, pages 1189--1232, 2001.

\bibitem{hand2001idiot}
David~J. Hand and Keming Yu.
\newblock Idiot's {B}ayes --- {N}ot so stupid after all?
\newblock {\em International Statistical Review}, 69(3):385--398, 2001.

\bibitem{howard2017mobilenets}
Andrew~G. Howard, Menglong Zhu, Bo~Chen, Dmitry Kalenichenko, Weijun Wang,
  Tobias Weyand, Marco Andreetto, and Hartwig Adam.
\newblock {MobileNets}: Efficient convolutional neural networks for mobile
  vision applications.
\newblock \url{https://arxiv.org/abs/1704.04861}, 2017.

\bibitem{huang2017densely}
Gao Huang, Zhuang Liu, Laurens Van Der~Maaten, and Kilian~Q. Weinberger.
\newblock Densely connected convolutional networks.
\newblock In {\em Proceedings of the IEEE conference on computer vision and
  pattern recognition}, pages 4700--4708, 2017.

\bibitem{jha2023label}
Rishi Jha, Jonathan Hayase, and Sewoong Oh.
\newblock Label poisoning is all you need.
\newblock {\em Advances in Neural Information Processing Systems},
  36:71029--71052, 2023.

\bibitem{ke2017lightgbm}
Guolin Ke, Qi~Meng, Thomas Finley, Taifeng Wang, Wei Chen, Weidong Ma, Qiwei
  Ye, and Tie-Yan Liu.
\newblock {LightGBM}: A highly efficient gradient boosting decision tree.
\newblock {\em Advances in Neural Information Processing Systems}, 30, 2017.

\bibitem{lecun1998gradient}
Yann LeCun, L{\'e}on Bottou, Yoshua Bengio, and Patrick Haffner.
\newblock Gradient-based learning applied to document recognition.
\newblock {\em Proceedings of the IEEE}, 86(11):2278--2324, 1998.

\bibitem{mehta2024natural}
Ritik Mehta, Olha Jure{\vvv{c}}kov{\'a}, and Mark Stamp.
\newblock A natural language processing approach to malware classification.
\newblock {\em Journal of Computer Virology and Hacking Techniques},
  20(1):173--184, 2024.

\bibitem{nappa2015malicia}
Antonio Nappa, M.~Zubair Rafique, and Juan Caballero.
\newblock The malicia dataset: identification and analysis of drive-by download
  operations.
\newblock {\em International Journal of Information Security}, 14:15--33, 2015.

\bibitem{paudice}
Andrea Paudice, Luis Mu{\~{n}}oz-Gonz{\'a}lez, and Emil~C. Lupu.
\newblock Label sanitization against label flipping poisoning attacks.
\newblock In Carlos Alzate et~al., editors, {\em ECML PKDD 2018 Workshops},
  pages 5--15, 2019.

\bibitem{rumelhart1986learning}
David~E. Rumelhart, Geoffrey~E. Hinton, and Ronald~J. Williams.
\newblock Learning internal representations by error propagation, parallel
  distributed processing, explorations in the microstructure of cognition.
\newblock {\em Biometrika}, 71:599--607, 1986.

\bibitem{Stamp_2022}
Mark Stamp.
\newblock {\em Introduction to Machine Learning with Applications in
  Information Security}.
\newblock Chapman and Hall/CRC, 2nd edition, 2022.

\bibitem{taheri2020defending}
Rahim Taheri, Reza Javidan, Mohammad Shojafar, Zahra Pooranian, Ali Miri, and
  Mauro Conti.
\newblock On defending against label flipping attacks on malware detection
  systems.
\newblock {\em Neural Computing and Applications}, 32:14781--14800, 2020.

\bibitem{xiao2015}
Huang Xiao, Battista Biggio, Blaine Nelson, Han Xiao, Claudia Eckert, and Fabio
  Roli.
\newblock Support vector machines under adversarial label contamination.
\newblock {\em Neurocomputing}, 160(C):53--62, 2015.

\end{thebibliography}
